\documentclass[letterpaper, 10 pt, conference]{ieeeconf}  

\IEEEoverridecommandlockouts                              

\usepackage{color}

\usepackage{graphicx}
\usepackage{amsmath}
\usepackage{amsthm}
\usepackage{amssymb}
\DeclareMathOperator*{\argmax}{argmax}
\DeclareMathOperator*{\argmin}{argmin}
\begin{document}
\theoremstyle{definition}
\newtheorem*{definition}{Problem}
\newtheorem*{remark}{Remark}

%
\title{Semantic SLAM with Autonomous Object-Level Data Association}
%
%
%

\author{Zhentian Qian,
        Kartik Patath, 
        Jie Fu, Jing Xiao
        
        \thanks{The authors are with the Department
 of Robotics Engineering, Worcester Polytechnic Institute, Worcester, MA,  USA, 01609  e-mail: zqian, kpatath, jfu2, jxiao2@wpi.edu}}

\maketitle

\begin{abstract}
\label{sec:Abstract}
It is often desirable to capture and map semantic information of an environment during simultaneous localization and mapping (SLAM). Such semantic information can enable a robot to better distinguish  places with similar low-level geometric and visual features and perform high-level tasks that use semantic information about objects to be manipulated  and environments to be navigated. 
While semantic SLAM has gained increasing attention, there is little research on semantic-level data association based on semantic objects, i.e., object-level data association. In this paper, we propose a novel object-level data association algorithm based on bag of words algorithm \cite{galvez2012bags}, formulated as a maximum weighted bipartite matching problem. With object-level data association solved, we develop a quadratic-programming-based semantic object initialization scheme using dual quadric and introduce additional constraints to improve the success rate of object initialization. The integrated  semantic-level SLAM system can achieve high-accuracy object-level data association and real-time semantic mapping as demonstrated in the experiments. The  online semantic map building and semantic-level localization capabilities facilitate semantic-level mapping and task planning in a priori unknown environment. 

\end{abstract}

\IEEEpeerreviewmaketitle

\section{Introduction}
\label{sec:Introduction}
%
%
%
%

In order for robots to interact intelligently with the real world and perform high level tasks, semantic information of its surroundings must be acquired. Traditional visual or visual-inertial simultaneous localization and mapping (SLAM) algorithms extract low-level geometric features such as corners, lines and surface patches from image sequence to build sparse or dense point cloud map. 
However, low-level geometric map is insufficient  for more sophisticated tasks, such as getting a book from a particular desk or carrying meal to a particular nightstand for a patient. It is often necessary to add semantic information into the map, which motivates the need for semantic SLAM.
While semantic SLAM has gained increasing attention in the literature (see Related Work), there is little research on semantic-level data association based on semantic objects. This paper introduces an integrated system for performing semantic-level SLAM, addressing the key problem of object-level data association. This endeavour is our first step towards a visual semantic SLAM algorithm which would tightly integrate geometric and semantic information.

The major contributions of our paper are the following:
\begin{itemize}
    \item an autonomous object-level data association algorithm utilizing both geometric and appearance information of the object;
    \item a novel object initialization scheme improving the success rate of object initialization;
    \item an integrated, keyframe-based, real-time semantic SLAM system without the use of pre-built object database. In other words, there is no need to survey the environment to build a database of object shape, size and appearance before running our semantic SLAM algorithm.
\end{itemize}


\subsection{Related Work}
\label{sec:related_work}
In order to bridge the gap between perception and action, the robotics community has taken a keen interest in semantic SLAM. The pioneering work of SLAM++  \cite{salas2013slam++} performs object-level SLAM using a depth camera. The main restriction being that an object database of both 3D shape and global description must be built in advance. Improving on \cite{salas2013slam++}, the work in \cite{galvez2016real} relies solely on monocular input and learns the scale of the map from object models. Nevertheless, the same restriction remains as a pre-built object database is still required. Refs. \cite{atanasov2018unifying, bowman2017probabilistic} resort to novel soft data association which in turn circumvents the need of having to assign incoming detected object to objects spawned in map. The drawback is that the object is still represented by 3D points instead of 3D shape. This representation limits the kind of interaction the robot can have with the physical world. Grasp operation, for example, may not be feasible with only point representation.

To address this issue of object representation, one method  \cite{nicholson2018quadricslam} used dual quadrics to capture the shape and pose of an object and developed a tailored SLAM backend for graph optimization. Dual quadric is a mathematically elegant solution as it can be compactly defined by nine continuous parameters and is also used for object representation in this paper. However, \cite{nicholson2018quadricslam} leaves the key problem of object-level data association unsolved. Another method  \cite{zhang2018semantic} integrates object detection module and RGB-D SLAM and further converts semantically augmented 3D point clouds to Octomap, which makes advanced missions such as grasp point selection possible. Although techniques including multi-threads processing are employed to speed up Octomap creation, their results generally take $50\sim200s$ to build the Octomap. Thus, this method falls short in real-time performance.


There are also works that directly add more semantics to the map generated from SLAM. The work in \cite{mccormac2017semanticfusion} uses a
SLAM system to provide correspondences from 2D frame into a 3D map. These correspondences allow the semantic prediction of a Convolutional Neural Network (CNN) from multiple viewpoints to be probabilistically fused into the map. However, the semantic predictions are only loosely added to the map and do not aid the task of SLAM. Similarly, the work in  \cite{kundu2014joint} also fuses 2D semantic segmentation and sparse 3D points from SLAM. This work is closely related to 3D reconstruction in computer vision as the 3D scene is represented in the form of voxels enhanced with semantic labels. Though the time for 3D reconstruction is not given in the paper, considering the size of the voxel representation and the time to perform optimization, real time performance is almost impossible. 



\section{System Overview}
\label{sec:System Overview}
\begin{figure}[htbp]
  \includegraphics[width=1\linewidth]{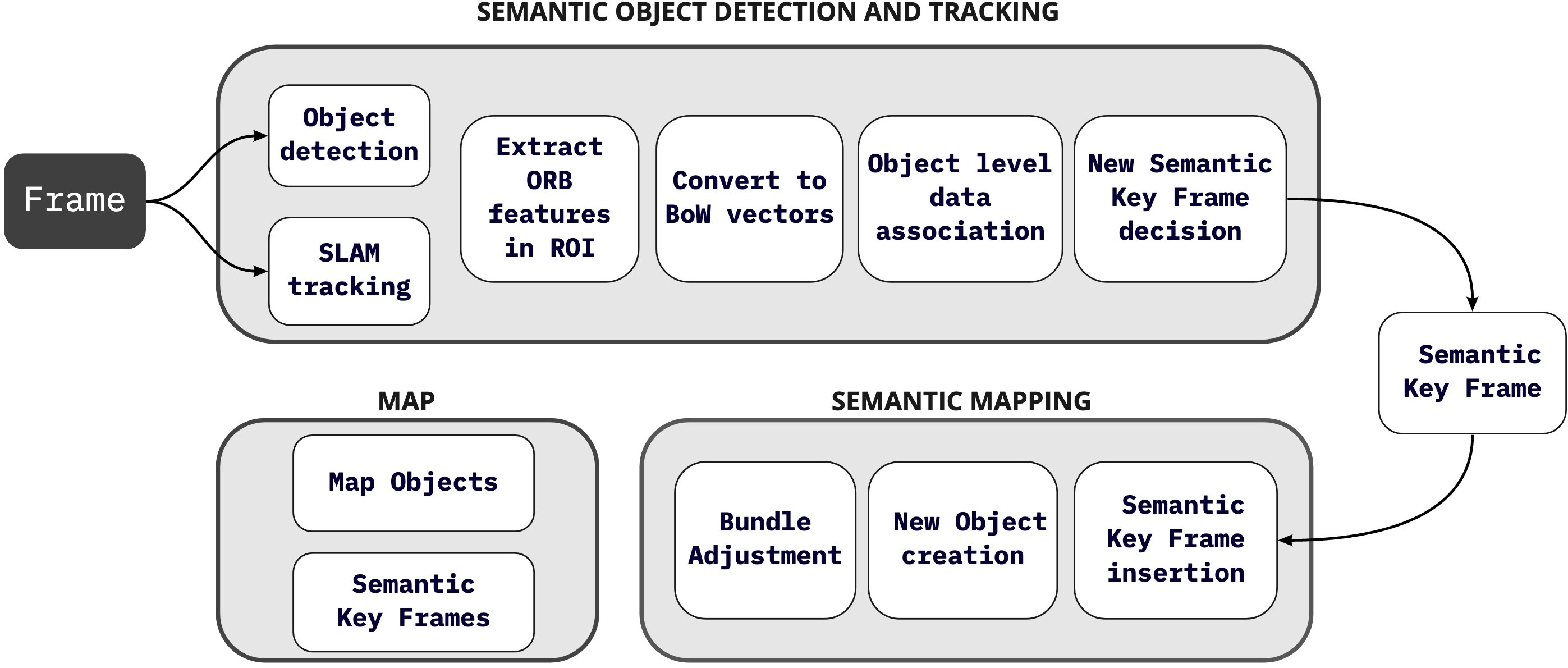}
  \caption{System overview, showing all the steps performed in semantic object detection and tracking thread and semantic mapping thread. The main components of the map are also shown.}
  \label{fig:system overview}
\end{figure}
 Figure~\ref{fig:system overview} presents an overview of our system composed of two threads: semantic object detection and tracking (Sec.~\ref{sec:Front End}) and semantic mapping (Sec.~\ref{sec:Mapping}). The map contains a set of map objects and semantic keyframes, as well as the correspondence between them.   A semantic keyframe stores the map objects it has observed in the frame image. At the same time, each map object records the sequence of semantic keyframes in which this object has been observed. Similar to \cite{nicholson2018quadricslam}, our semantic SLAM employs semantic objects as landmarks and represents them as dual quadrics.  The attributes of map objects and semantic keyframes are described in the Appendix.  Next, we provide more details about the algorithms used in these two threads.

\section{Semantic Object Detection and Tracking}
\label{sec:Front End}
In  this  section,  we describe the semantic object detection and tracking thread. The key component of this thread is object-level data association---an essential step for robust semantic SLAM.  Before we could perform data association, we first perform object detection and extract ORB features in the region of interest (ROI) where objects are detected. These features are subsequently converted to Bag of Words (BoW) \cite{galvez2012bags} vectors to describe the detected objects. Finally, the data association step is performed to match the semantic measurements generated by object detector to map objects.

\subsection{Object Detection and Conversion to BoW vectors}
Our algorithm uses an ROS implementation \cite{bjelonicYolo2018} of YOLOv3 \cite{redmon2018yolov3} as real-time object detector. For every processed RGB or grey image $I_t $, YOLOv3 outputs a set of semantic measurements $\mathcal{S}_t = \{z_k\}$, where each semantic measurement $z_k = \{b_k, c_k, s_k\}$ is a collection of three attributes, i.e., bounding box $b_k$, object class $c_k$, and detection score $s_k$. We discard semantic measurements with bounding boxes smaller than a certain threshold as we deemed them to hold too little image content to be robustly tracked.

The bag of words algorithm \cite{galvez2012bags}  converts the 2D image content inside a bounding box into a 1D BoW vector describing the appearance of the detected objects. The BoW vectors are compact to store and easy to compare. Each entry of the BoW vector is a weighted occurrence count of a particular visual word, i.e., a discretized ORB descriptor space. ORB features are required in the conversion. For each ROI bounded by a bounding box, FAST corners are extracted at a consistent density and the number of corners extracted per ROI varies according to the area of the ROI. For ROI that is textureless or has low contrast and holds few corners, its corresponding semantic measurement and corners extracted in this ROI are discarded. The ORB descriptors are then computed for the remaining FAST corners.  

In the next step, we employ visual vocabulary \cite{galvez2012bags}, which is the discretization of the descriptor space, to convert the ORB features extracted in each bounding box to BoW vectors. It is structured as a tree with binary nodes which are created by $k$-medians clustering of training ORB descriptors. The leaves of the tree are the words of the visual vocabulary, weighted with the term frequency - inverse document frequency (tf-idf) according to their relevance in the training corpus. Words with fewer occurrences in the training images are deemed more discriminative and given a higher weight. Instead of using a single vocabulary, we create vocabularies for every object class in our implementation. Our intention is to ensure that each vocabulary is uniquely suited to distinguish the image content in bounding boxes of a particular object class. Each vocabulary is built offline with the ORB descriptors extracted within the ground truth bounding boxes from the COCO dataset images \cite{lin2014microsoft}. With $k = 5$ branches and $L = 5$ depth levels, each vocabulary has 3125 words. 

When ORB features are extracted from a bounding box of certain object class, we use the matching visual vocabulary to convert them into a BoW vector. The conversion process is described as follows: For all the given ORB features, their descriptor vectors traverse the vocabulary tree from the root to the leaves, selecting at each level the nodes which minimizes the Hamming distance \cite{waggener1995pulse}. The BoW vector is simply the weighted occurrence counts of the words, i.e., leaves. The BoW vector provides a way for us to quantify the similarity between the semantic measurements and the map objects based on their appearance.

\subsection{Object-Level Data Association} 
We propose an object-level data association scheme to match the incoming semantic measurements from YOLOv3 to registered objects in the map. Our method is a two-step frame-to-map matching, described as follows.

\begin{figure}[htbp]
  \centering
  \includegraphics[width=.7\linewidth]{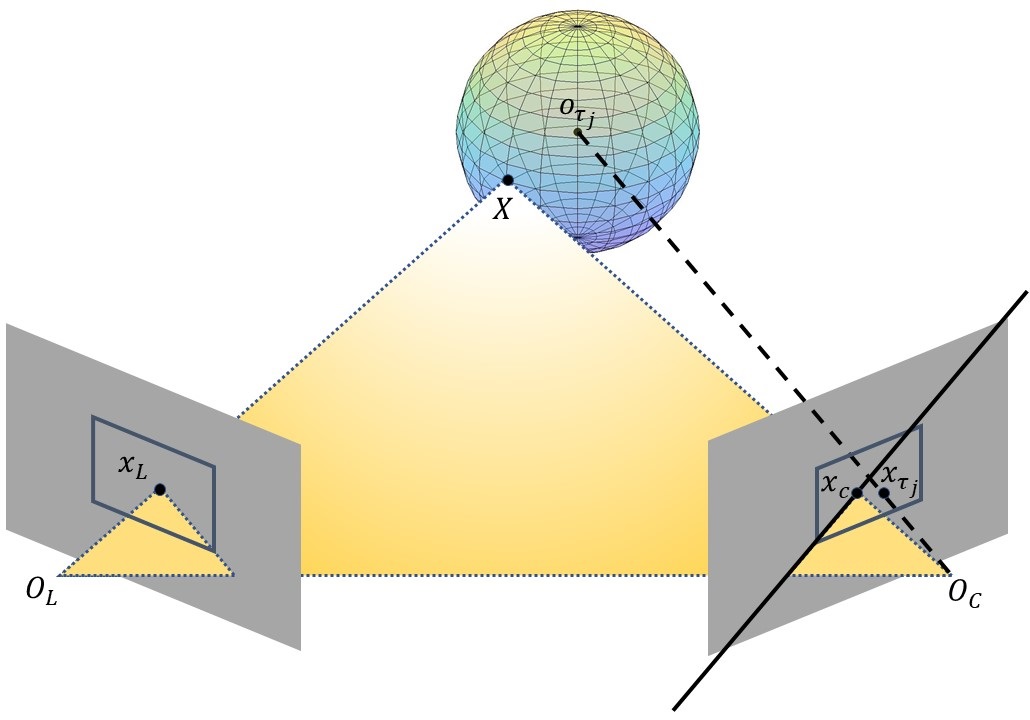}
  \caption{The geometric requirements imposed on matching object candidates}
  \label{fig:DA geometry}
\end{figure}

The first step is to determine the set of matching object candidates $\mathcal{T}^{ca}$ for a semantic measurement $z_k \in \mathcal{S}_t$. Each map object $\tau_j$ contains its class label. For a semantic measurement $z_k= \{b_k, c_k, s_k\}$, only objects with class label $c_k$ are considered. Apart from the requirement on object class, we introduce geometric checks to rule out objects with similar appearances but incorrect locations. Consider two cases:

\smallskip
\noindent {\bf Case 1}: The map object $\tau_j$ has quadric representation. In this case, the projection of the object quadric center should reside inside the bounding box of the semantic measurement $z_k$, as shown on the right frame of Fig. \ref{fig:DA geometry}.

\smallskip
\noindent {\bf Case 2}: The map object $\tau_j$ does not have quadric representation. In this case, we assume that the 3D point corresponding to the center of the bounding box of the latest observation of $\tau_j$ is projected inside the bounding box of $z_k$. Based on this assumption, the corresponding epipolar line should go through the bounding box of $z_k$. The epipolar line is represented by the real line on the right frame of Fig. \ref{fig:DA geometry}.

Once we have determined the set of matching object candidates $\mathcal{T}^{ca}$, we proceed to the second step of data association. This step performs data association based on object appearance. A data association score is calculated between the semantic measurement and every potential matching object. Note that the assignment of semantic measurements to map objects is coupled, since the assignment of one semantic measurement to a particular map object means that map object can no longer be considered as candidate for other measurements. Our system solves the assignment of all semantic measurements by trying to maximize the sum of data association score. The details are given below.

For every map object $\tau_j \in \mathcal{T}^{ca}$, our system goes though semantic keyframes $K_i \in \mathcal{K}^{j}$, where $\mathcal{K}^{j}$ stands for the set of semantic keyframes that share  observation of the same object $\tau_j$. A $L_1$-score $s(\mathbf{v}_1, \mathbf{v}_2) = 1 - 0.5\big\lvert \mathbf{v}_1/ \lvert \mathbf{v}_1 \rvert -\mathbf{v}_2/\lvert \mathbf{v}_2 \rvert \big\rvert$ is calculated between the BoW vector stored in the semantic keyframe $K_i$ and the BoW vector corresponding to the bounding box of the semantic measurement $z_k$. The maximum score among $\mathcal{K}^{j}$ is considered the data association score $c_{kj}$ between the semantic measurement $z_k$ and the map object $\tau_j$. Next, we formulate the   object-level data association  as a maximum weighted bipartite matching problem (assignment problem). First, we introduce a set of Boolean decision variables: For each semantic measurement $z_k$ and a map object $\tau_j$, let
\begin{equation}
    x_{kj} = \begin{cases}
    1, & \text{if } z_{k} \text{ is assigned to } \tau_j,\\
    0, & \text{otherwise.}
    \end{cases}
\end{equation}
 And the assignment problem is formulated as an integer-programming problem as follows:
\begin{align}
\label{eqn:AP}
    \argmax_{x_{kj}} & \sum_{\{k:z_k \in \mathcal{S}_t^{c}\}}\sum_{\{j:\tau_j \in \mathcal{T}^{ca}\}} c_{kj}x_{kj} \\
    \textrm{s.t.} &     \sum_{\{j:\tau_j \in \mathcal{T}^{ca}\}} x_{kj} \leq 1\text{,\quad} \sum_{\{k:z_k \in \mathcal{S}_t^{c}\}} x_{kj} \leq 1.
    \label{eq:constraint} 
\end{align}
where  $\mathcal{S}_t^c$ is the set of semantic measurements taken on image $I_t$ with class $c$. The constraints \eqref{eq:constraint} mean that (a) each semantic measurement can only be assigned to at most one map object and some semantic measurements cannot be assigned to map objects because they are either from new objects that have not yet been observed or due to false detections;
and (b) each map object could only be associated with at most one semantic measurement.

The problem defined in (\ref{eqn:AP}) can be solved using a cost-scaling push-relabel algorithm \cite{goldberg1995efficient, kennedy1995solving, burkard2012assignment}, readily implemented in \cite{ortools}. This algorithm has a time complexity of $\mathcal{O}(\sqrt{n}m\log{(nC)})$, where $n$ is the number of nodes in the bipartite graph, $m$ is the number of edges in the bipartite graph, and $C$ is the largest edge cost (all edge costs need to be converted to integers). 
\begin{remark}
YOLOv3 could sometimes assign different class labels to the same object in some corner cases. Performing data association only considering objects with the same class label would subsequently lead to wrong matches. Future work would calculate a matching score between the class of the map objects and the class of the semantic measurement instead of simply ruling out map objects of different classes.
\end{remark}

\section{Semantic Mapping}
\label{sec:Mapping}
In this section, we describe the semantic mapping thread, which is performed on every new semantic keyframe $K_i$. The semantic keyframe is generated by the semantic object detection and tracking thread every $T$ image frames. When a  new semantic keyframe is added, we update the map database to include this keyframe as well as the relations between map objects and semantic keyframes.   



\subsection{New Map Object Creation and Initialization}
\label{ssec:Initialization}
In this section, we propose a novel object initialization scheme. It is observed that sometimes the object initialized based on \cite{nicholson2018quadricslam} is behind or intersects with the camera principal plane. We address this problem with a new object initialization scheme which can increase the success rate of object initialization.

Unlike in geometric SLAM where a map point is created and initialized at the same time, in our approach, a new object
$\tau_j$ is created and registered in the map if it is observed the first time in the current semantic keyframe. 
Upon creation, the class of $\tau_j$ is determined, the current semantic keyframe observing $\tau_j$ is stored, and the number of observations is set to one. However, some attributes of $\tau_j$ cannot be initialized by one observation, i.e., the shape, rotation, and translation (refer to Appendix). Only after a sufficient number of observations have been made on $\tau_j$, its initialization procedure is performed. The initial shape and pose for the quadric representation of $\tau_j$ is calculated from semantic measurements $z_k$ stored in a sequence of semantic keyframes $K_i \in \mathcal{K}^{j}$.

A quadric,  in its dual form, is represented by the locus of all planes tangent to the quadric. A tangent plane $\Pi$ satisfies:
\begin{equation}
    \Pi^T\mathbf{Q}^*\Pi=0
    \label{eq:quadric def}
\end{equation}
where ${\bf Q}^*$ is a $4 \times 4$ symmetric matrix defining a quadric and only defined up to scale, and  
$\Pi$ is a $4\times 1$ vector corresponding to the coefficients of general form of the plane equation. Note that coefficients of plane equations are the parametrization of all planes mentioned below.

We can represent a generic dual quadric with a 9-vector $\hat{q} = [\hat{q}_1, \cdots, \hat{q}_{9}]$ where each element corresponds to one of the first nine independent elements of symmetric matrix $\mathbf{Q}^{*}$. The last element of $\mathbf{Q}^{*}$ is set to $-1$ to define the scale of $\mathbf{Q}^{*}$ to be $1$. The quadric representation of map object $\tau_j$ can be initialized with quadratic programming to fit its defining equation \eqref{eq:quadric def}. Expanding \eqref{eq:quadric def} leads to the following linear equation:
\begin{multline}
\big[\Pi(1)^2, 2\Pi(1)\Pi(2), 2\Pi(1)\Pi(3), 2\Pi(1)\Pi(4), \Pi(2)^2, \\2\Pi(2)\Pi(3), 2\Pi(2)\Pi(4), 
\Pi(3)^2, 2\Pi(3)\Pi(4), \Pi(4)^2 \big] \\
\cdot
[\hat{q}_1, \hat{q}_2,\hat{q}_3, \hat{q}_4, \hat{q}_5, \hat{q}_6, \hat{q}_7, \hat{q}_8, \hat{q}_9, -1]^T = 0.
\label{eq:linsys}
\end{multline}

For each semantic measurement $z_k$ taken on semantic keyframe $K_i \subset \mathcal{K}^{j}$ and associated with $\tau_j$, we can generate four tangent planes to the quadric of the object $\tau_j$ based on its bounding box:
\begin{equation}
\begin{split}
     \Pi_{k,xmin}^T = [1, 0, -x_{k,min}]\mathbf{P}_i  &\text{,\;} \Pi_{k,xmax}^T = [1, 0, -x_{k,max}] \mathbf{P}_i\\
     \Pi_{k,ymin}^T = [0, 1, -y_{k,min}]\mathbf{P}_i  &\text{,\;} \Pi_{k,ymax}^T = [0, 1, -y_{k,max}] \mathbf{P}_i.
\end{split}
\label{eq:tangent plane}
\end{equation}
Here the camera projection matrix $\mathbf{P}_i$ is calculated from the camera pose $\mathbf{T}^i_w$ recorded in $K_i$. The camera pose $\mathbf{T}^i_w$ is initialized from the odometry measurements.

By collecting all tangent planes generated from the semantic measurements that are associated to object $\tau_j$ and substitute them into \eqref{eq:linsys},
we obtain a linear system of the form $\mathbf{A}_j\begin{bmatrix}\hat{q}\\ -1\end{bmatrix} = 0$ with $\mathbf{A}_j$ containing the coefficients of all tangent planes to the quadric of map object $\tau_j$ as in \eqref{eq:linsys}. For the quadratic programming based initialization process, the objective function is the sum of square errors of the residual on the left hand side of \eqref{eq:linsys}. Formally,
\begin{equation}
    \min_{\hat{q}} \frac{1}{2}\hat{q}^T\mathbf{H}\hat{q}+f^T\hat{q}.
\label{eq:objfun}
\end{equation}

Let
\begin{equation}
    \mathbf{B} = \mathbf{A}_j^T\mathbf{A}_j = \begin{bmatrix}
    \mathbf{B}_{99} & \mathbf{B}_{91} \\
    \mathbf{B}_{19} & \mathbf{B}_{11}
    \end{bmatrix}
\end{equation}
where $\mathbf{B}_{ij}$ is a submatrix of $i$ rows and $j$ columns. Then $\mathbf{H} = \mathbf{B}_{99}$ and $f = \mathbf{B}_{91}$. 

To improve the success rate of object initialization, we propose to include new constraints preventing the objects from being incorrectly initialized. The first set of constraints specify that the initialized quadric should be in front of the camera:
\begin{equation}
(o_{\tau_j} - o_{K_i})\cdot z_{K_i} \geq 0, \forall K_i \in \mathcal{K}^{j}
\label{eq:cs1}
\end{equation}
where $o_{K_i}$ is the camera center of semantic keyframe $K_i$, $z_{K_i}$ is the optical axis of the camera of $K_i$, $o_{\tau_j}$ is the quadric center of object $\tau_j$ and $o_{\tau_j} = -
\begin{bmatrix}
\hat{q}_4 & \hat{q}_7 & \hat{q}_9  
\end{bmatrix}^T$. Both $o_{K_i}$ and $z_{K_i}$ can be extracted from the camera pose $\mathbf{T}^i_w$ of $K_i$. 

The second set of constraints specify that the principal plane of the camera should not intersect with the quadric:
\begin{equation}
    \Pi_{K_i}^T \mathbf{Q}^*_{(\hat{q}_j)} \Pi_{K_i} \leq 0, \forall K_i \in \mathcal{K}^{j}
    \label{eq:cs2}
\end{equation}
where $\Pi_{K_i} = \begin{bmatrix}
z_{K_i} \\ -z_{K_i} \cdot o_{K_i}
\end{bmatrix}$ is the principal plane of $K_i$. It is not be be confused with the tangent planes defined in \eqref{eq:tangent plane}. 

The third set of constraints specify that the projection of $o_{\tau_j}$ should remain inside the bounding box. For each semantic measurement $z_k$ taken on semantic keyframe $K_i \in \mathcal{K}^{j}$ and associated with $\tau_j$, we have
\begin{equation}
   x_{k,min} \leq u \leq x_{k,max} \text{,\quad} y_{k,min} \leq v \leq y_{k,max}
    \label{eq:cs3}
\end{equation}
where $[u,v,1]^T = P_io_{\tau_j}$ is the projection of $o_{\tau_j}$ on semantic keyframe $K_i$.

Constraints \eqref{eq:cs1}, \eqref{eq:cs2} and \eqref{eq:cs3} are linear. A quadratic programming problem with linear constraints can be readily solved using convex optimization tools such as CGAL \cite{cgal:eb-20b}. 

The solution of the quadratic programming  represents a generic quadric surface, not necessarily an ellipsoid; we therefore constrain each quadric to be an ellipsoid by extracting the quadric rotation, translation and shape as shown in \eqref{eq:ellipsoid}, using methods introduced in \cite{rubino20173d}:
\begin{equation}
    \mathbf{Q}^{*} =\begin{bmatrix}
                    \mathbf{R}\operatorname{diag}(a^2, b^2, c^2)\mathbf{R}^T -tt^T & t \\
                    t^T     & -1
                \end{bmatrix}
\label{eq:ellipsoid}
\end{equation}
where $t = [t_1, t_2, t_3]^T$ is the translation vector, $\mathbf{R}$ is the $3\times 3$ rotation matrix and $a, b, c$ are the semi-axes of the ellipsoid.





\begin{figure}[htbp]
  \centering
  \includegraphics[width=.64\linewidth]{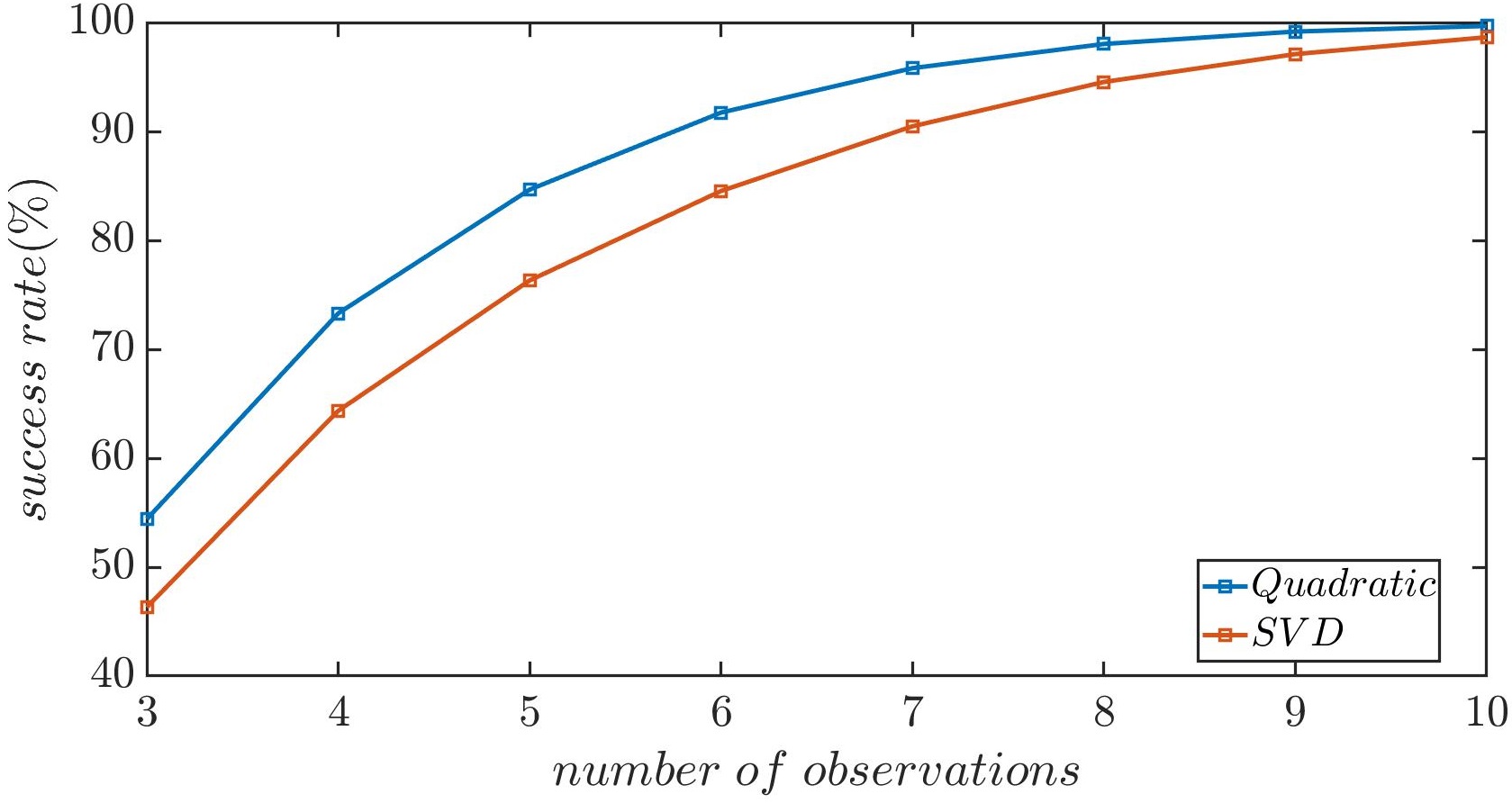}
  \caption{The initialization success rate}
  \label{fig:initialization success rate}
\end{figure}

Hence, we initialize all map objects by solving the quadratic programming problem \eqref{eq:objfun} over the complete set of detections for each map object, and constrain the estimated quadrics to be ellipsoid. The constrained quadrics are deemed failure and discarded if they fail to satisfy constraints \eqref{eq:cs1}, \eqref{eq:cs2} and \eqref{eq:cs3} or have an average re-projection error greater than 100 pixels. Note that our initialization method (marked by 'Quadratic') yields a higher success rate than the method originally proposed in \cite{nicholson2018quadricslam} (marked by 'SVD'), as shown in Fig. \ref{fig:initialization success rate}. Another observation is that quadric initialization has a higher success rate if more observations are utilized, hence in practice, we require a minimum number of ten observations to ensure robust initialization.

\subsection{Bundle Adjustment}
\label{ssec:BA}
The object initialization process only utilizes the first few observations and the generated quadrics are still sub-optimal. In order to take the full advantage of all the observations, we perform a joint optimization of map objects and camera poses of semantic keyframes in bundle adjustment.

Our data association pipeline would produce few errors, as demonstrated in the experiment section. With accurate data association, the maximum a posterior (MAP) estimation of map objects and camera poses of semantic keyframes can be solved to make improvements on the initialization results by maximizing the product of factors:
\begin{equation}
\begin{split}
     \mathcal{X}^*, \mathcal{T}^*  = & \argmax_{\mathcal{X}, \mathcal{T}} 
      \underbrace{\prod_{i} p(u_{i+1}\;|\;x_{i+1}, x_i)}_{\text{Odometry}} \cdot \underbrace{\prod_{i} p(x_i)}_{\text{Pose Prior}}\cdot \\
     & \underbrace{\prod_{j}\prod_{\{i:K_i \in \mathcal{K}^{j}\}} p(z_i^j\;|\;x_i, \tau_j)}_{\text{Semantic}} \cdot\underbrace{\prod_{j}p(\tau_j)}_{\text{Object Prior}}
     \label{eq:factor graph}
\end{split}
\end{equation}
where $\mathcal{T} = \{\tau_j\}$ is the set of map objects, $\mathcal{X} = \{x_i = T^i_w\}$ is the set of camera poses of semantic keyframes, $u_i$ is the odometry measurement and $z_i^j$ is the semantic measurement of object $\tau_j$ on semantic keyframe $K_i$. Further, assuming Gaussian measurement and process models and uniform distribution of object and pose, by taking the negative log on the objective function, \eqref{eq:factor graph} can be rewritten as a nonlinear least-squares problem:
\begin{equation}
\begin{split}
      \mathcal{X}^*, \mathcal{T}^*  &  = \argmin_{\mathcal{X}, \mathcal{T}} 
      \sum_{i} \lVert h_o(x_{i+1}, x_i) - u_{i+1}\rVert^2_{\Sigma_u}  \\
     & + \sum_{j}\sum_{\{i:K_i \in \mathcal{K}^{j}\}} \lVert h_s(x_i, \tau_j) - z_i^j\rVert^2_{\Sigma_z} 
\end{split}
\label{eq:nls}
\end{equation}
where $\lVert\cdot\rVert_\Sigma$ is the Mahalanobis norm, $\Sigma_u$ and $\Sigma_z$ are the covariance matrices of odometry measurements and semantic measurements respectively, $h_o$ and $h_s$ are the sensor models of odometry and semantic measurements respectively. While $h_o$ is already known, $h_s$ needs to be established. Since we perform data association for objects and semantic measurements  with  the  same  class  label, the predicted object class is always aligned with the measurements. On the other hand, there would be discrepancy between predicted bounding box and the bounding box from semantic measurement. The predicted bounding box is calculated from the dual conic projection from object $\tau_j$ on semantic keyframe $K_i$:
\begin{equation}
    C_{ij}^* = P_iQ_{\tau_j}^*P_i^T
\end{equation}
where $C_{ij}^*$ is a $3\times 3$ matrix defining the dual conic projection, $P_i$ is the projection matrix. The predicted bounding box $h_s(x_i, \tau_j)$ is set to be the smallest bounding box containing the part of the conic on the image. 


With the established sensor model, the problem described in \eqref{eq:nls} can be readily solved using modern libraries such as g2o \cite{kummerle2011g}. If a semantic keyframe is inserted when bundle adjustment is busy, a signal is sent to stop bundle adjustment, so that the semantic mapping thread can process the new semantic keyframe as soon as possible.


\section{Experimental Evaluation}
\label{sec:Experimental Evaluation}

\subsection{Data Association}
We evaluate the data association performance on the TUM RGB-D fr1\_xyz sequence \cite{sturm11rss-rgbd}. The original sequence is appended with semantic measurements from YOLOv3. The ground truth object IDs are then manually labeled for those semantic measurements.

\begin{figure}[htbp]
  \centering
  \includegraphics[width=\linewidth]{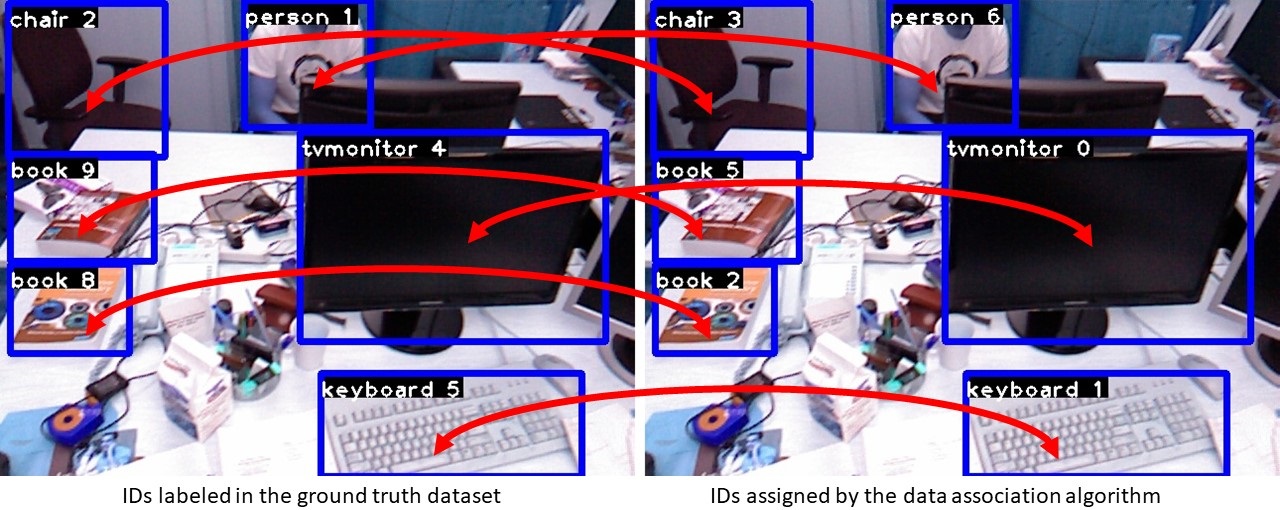}
  \caption{The correspondences between the IDs labeled in the ground truth dataset and the IDs assigned by the data association algorithm are found by solving the maximum weighted bipartite matching problem.}
  \label{fig:DA results}
\end{figure}

First, we clarify how accuracy is evaluated in our experiment. The key idea is to find the correspondences between the object IDs assigned to semantic measurements in ground truth dataset and the IDs assigned by our data association algorithm, as shown in Fig. \ref{fig:DA results}. This task is formulated and solved as a maximum weighted bipartite matching problem. The complete bipartite graph has two disjoint vertices sets $U$ and $V$. The IDs labeled in the ground truth dataset form the vertices set $U$. The IDs assigned in the data association algorithm form the vertices set $V$. The reward function $r(i, j)$ on a particular edge $(i,j) \in U\times V$ is the number of semantic measurements assigned ID $i \in U$ in the ground truth dataset and ID $j \in V$ in the data  association algorithm respectively. For example, if our dataset only consists of the one image shown in Fig. \ref{fig:DA results}, then we would have $r(8, 2) =1$ and $r(8, 5) = 0$ because book object 8 on the left and book object 5 on the right do not share the same bounding box.
Assume IDs assigned by our data association algorithm are identical to the IDs in ground truth dataset on every frame, then the total reward we obtain is simply the number of semantic measurements assigned IDs in both the ground truth dataset and the data association algorithm. This reward, denoted by $r_{max}$, is the maximum possible reward. By solving the maximum weighted bipartite matching problem, we obtain a particular matching. The sum of rewards of all edges inside that matching is considered the rewards obtained by our data association algorithm $r_{da}$. The ratio $r_{da}/r_{max}$ is considered the accuracy of our data association algorithm.

The results show that the accuracy of our data association algorithm is $790/857 \approx 92.19\%$. The errors can be put into two categories: (I) 60 mismatch cases are due to YOLOv3 assign different object classes to the same object. Since our data association algorithm assumes that semantic measurements with different object classes correspond to different map objects, errors can occur. (II) 9 mismatch cases are caused by errors occurred inside our data association pipeline. Those errors tend to happen for objects with textureless surface, such as the monitor shown in Fig. \ref{fig:DA results}.

\subsection{Results of Semantic Mapping}
\begin{figure}[htbp]
  \centering
  \includegraphics[width=.71\linewidth]{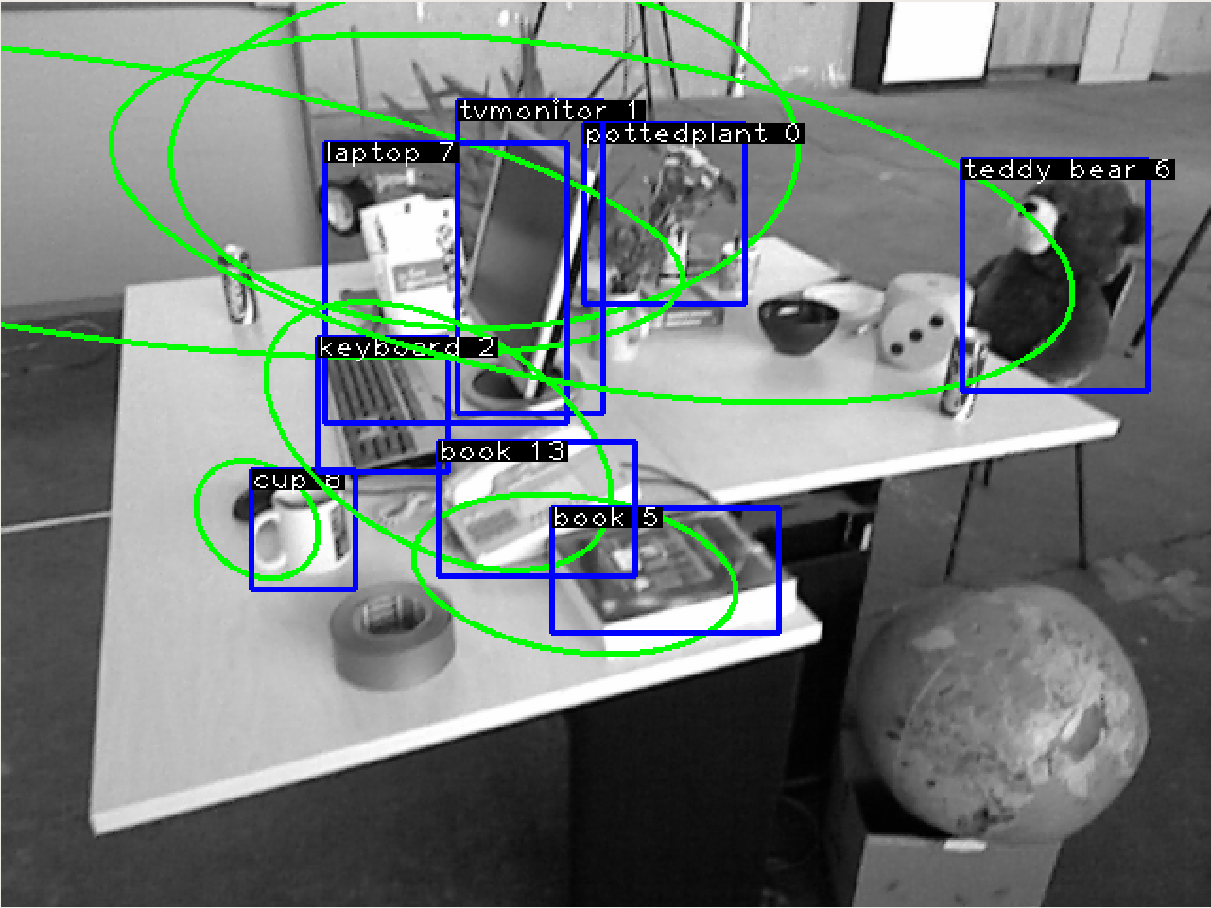}
  \caption{Results from only initialized map objects (in green).}
  \label{fig:initialization restuls}
\end{figure}

\begin{figure}[htbp]
  \centering
  \includegraphics[width=.71\linewidth]{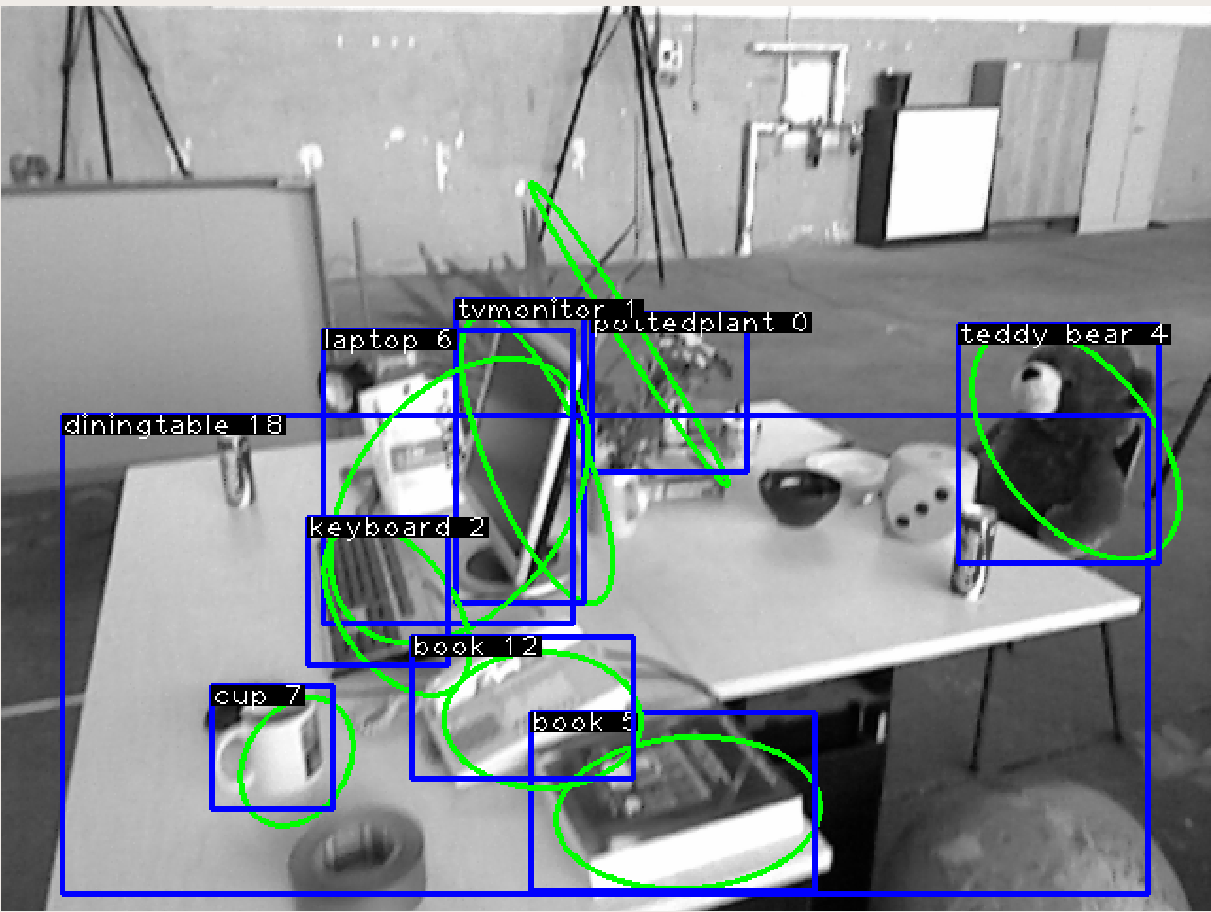}
  \caption{Results after performing bundle adjustment (in green).}
  \label{fig:BA restuls}
\end{figure}

Evaluated on the TUM RGB-D fr2\_desk sequence, the results of semantic mapping are shown in Fig. \ref{fig:initialization restuls} and \ref{fig:BA restuls}. Fig. \ref{fig:initialization restuls} shows the projected quadrics of map objects (in green) generated by our proposed initialization scheme on the image plane. The blue bounding boxes were results from YOLOv3. For initialization, only the first few observations are utilized. Hence, when observing from frames whose parallaxes with respect to the initialization frames are significant, we can see that the initialization results are far from ideal. To quantify the accuracy of mapping, we calculate the reprojection error \cite{hartley2003multiple} for all objects. The average reprojection error per object per frame is 87 pixels.

The bundle adjustment described in \ref{ssec:BA} takes on average $136 ms$ to finish. Based on the current semantic keyframe rate (one out of every $T$=$4$ image frames is chosen as the semantic keyframe), we can process incoming frames at a $30Hz$ rate, which is reasonable to perform in real-time for most mapping tasks. Fig. \ref{fig:BA restuls} shows the results after performing the bundle adjustment. We can see that improvements have been made upon the initialization results, and the quadric projections on the image tightly fit the bounding boxes with the exception of the potted plant 0. The quadric representation of this object is optimized to a low-volume ellipsoid. This problem will be addressed in future work by incorporating object shape prior to bundle adjustment. The average reprojection error per object per frame decreases to 29 pixels  after performing bundle adjustment. The results suggest that the optimized quadrics in 3D space are reasonable  representations of the actual objects.

\section{Conclusions}
\label{sec:Conclusion}
In this paper, we have developed an integrated,  keyframe-based  semantic  SLAM  system without the use of pre-built object database. We have addressed the key problem of object-level data association utilizing both geometric and appearance information of map objects. We have also proposed a novel object initialization scheme to boost the success rate of object initialization and effectiveness of data association. The experimental results show that our data association algorithm has an accuracy of $92.19\%$ and that the generated semantic map is a reasonable presentation of the objects in the environment. 

Albeit the good results on data association and mapping, we have observed an increase in tracking errors comparing to ORB-SLAM2 \cite{mur2017orb}. Evaluated on the TUM RGB-D fr2\_desk sequence, the error increases from $2.0cm$ to $5.9cm$ if only high-level semantic information and odometry measurements are used, which suggests the need for combining high-level semantic information and low-level geometric information in cases when high-accuracy in tracking performance is needed. This will be one further step of our work. We will also consider incorporating semantic SLAM and motion and task planning in the future. 

\appendix
\paragraph*{The attributes of map objects}
In each map object $\tau_j$, the following attributes are stored:
\begin{itemize}
  \item  Quadric shape represented  by three semi-axes: $a$, $b$, $c$.
  \item Rotation matrix of the quadric $\mathbf{R}$.
  \item Translation of the quadric $o_{\tau_j} = t$.
  \item The class label of the object.
  \item The set of semantic keyframes $\mathcal{K}^{j}$ that can observe this object.
  \item The number of observations.
\end{itemize}

\paragraph*{The attributes of semantic keyframes}
Each semantic keyframe $K_i$ stores the following information:
\begin{itemize}
  \item The camera pose $\mathbf{T}^i_w$ representing the world frame with respect to the camera frame, in terms of a homogeneous transformation matrix.
  \item The camera intrinsics, including focal length and principal point.
  \item Objects observed in this semantic keyframe.
  \item The set of semantic measurements taken at this frame, associated or not to a map object.
  \item BoW vectors for the ROI defined by bounding boxes, used to describe observed map objects.
\end{itemize}
 
\bibliographystyle{IEEEtran}
 \bibliography{bibtex/bib/refs.bib}

\end{document}